\documentclass{article}
\usepackage{iclr2027_conference,times}
\usepackage[T1]{fontenc}
\usepackage{amsmath,amssymb,booktabs,tabularx,graphicx,microtype,xcolor}
\usepackage{algorithm,algpseudocode}
\usepackage{tikz}
\usetikzlibrary{arrows.meta,positioning,calc}
\usepackage{hyperref}
\usepackage{url}
\hypersetup{colorlinks=true,linkcolor=blue!45!black,citecolor=blue!45!black,urlcolor=blue!45!black}
\definecolor{marcblue}{HTML}{245B78}
\definecolor{marcgreen}{HTML}{32786B}
\definecolor{marcorange}{HTML}{A66A2C}
\newcommand{\method}{\textsc{MARC}}

\title{MARC: Mechanism-Aware Reflective Correction\\for Articulated Object Code Generation}
\iclrfinalcopy
\author{%
Xiong Yang$^{1,2,*}$\hfill
Jinkun Hao$^{1,*}$\hfill
Zijian Xiao$^{1,2}$\hfill
Zipeng Ye$^{2,\dagger}$\hfill
Yuchen Xie$^{2}$\hfill
Ran Yi$^{1,\dagger,\ddagger}$%
}
\makeatletter
\renewcommand{\@maketitle}{%
\begin{center}
{\fontsize{17}{20}\selectfont\bfseries\@title\par}
\vspace{10pt}
{\fontsize{10.5}{13}\selectfont\makebox[\textwidth][s]{\@author}\par}
\vspace{6pt}
{\normalsize $^{1}$Shanghai Jiao Tong University\qquad $^{2}$Meituan\par}
\vspace{5pt}
{\footnotesize $^{*}$Equal contribution.\quad $^{\dagger}$Project leader.\quad $^{\ddagger}$Corresponding author.\par}
\end{center}
\vspace{2pt}%
}
\makeatother
\renewenvironment{abstract}{%
\par\vspace{4pt}\begin{center}\bfseries Abstract\end{center}\vspace{-4pt}\noindent
}{\par\vspace{8pt}}

\hypersetup{pdftitle={MARC: Mechanism-Aware Reflective Correction for Articulated Object Code Generation},pdfauthor={Xiong Yang; Jinkun Hao; Zijian Xiao; Zipeng Ye; Yuchen Xie; Ran Yi}}
\begin{document}
\maketitle
\fancyhead{}
\renewcommand{\headrulewidth}{0pt}

\begin{figure}[H]
\centering
\includegraphics[width=\linewidth]{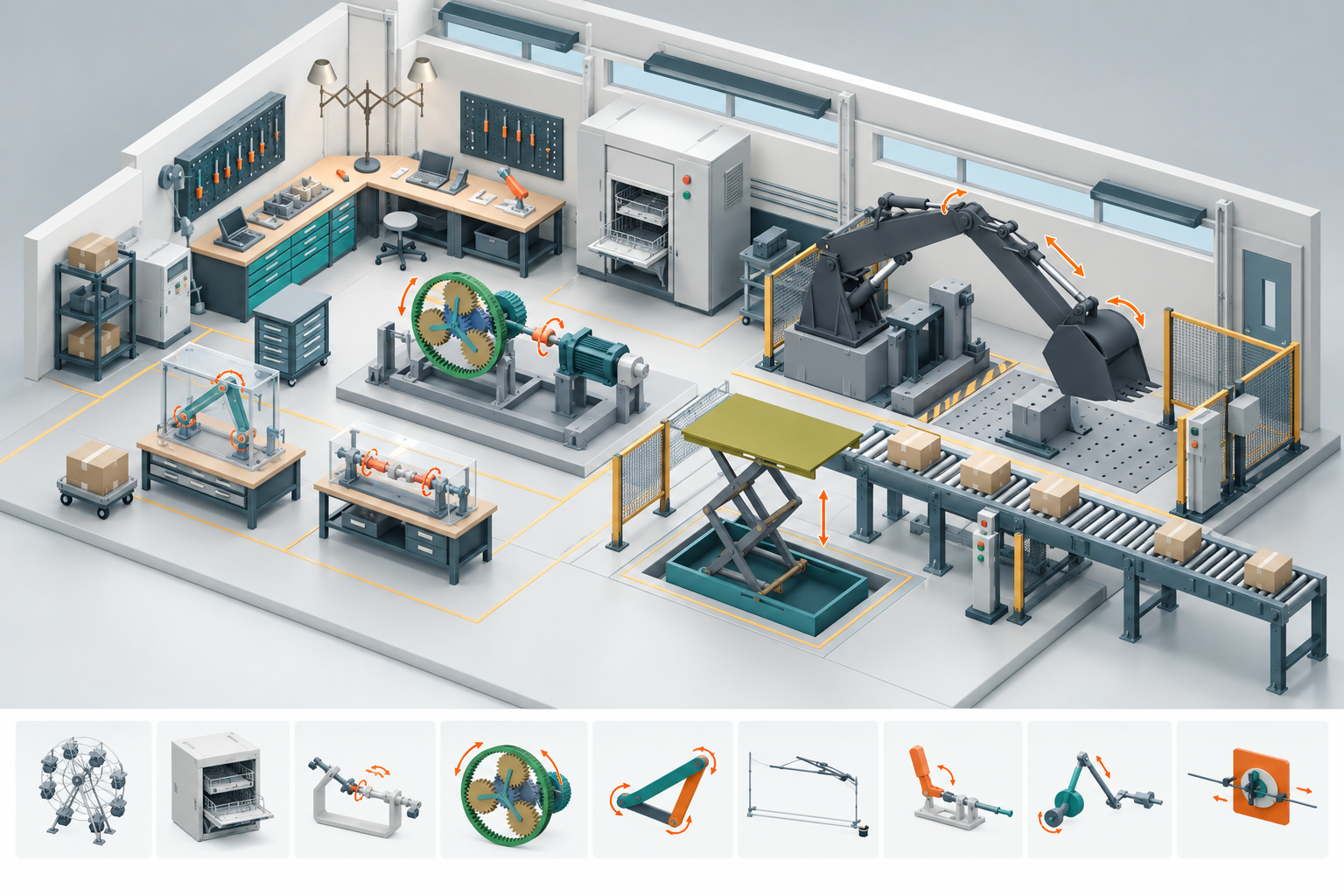}
\caption{Articulated objects generated by \method{}, including linkages, transmissions, and everyday mechanisms. The workshop view summarizes the range of generated objects; the display brings these mechanisms into a single scene.}
\label{fig:teaser}
\end{figure}

\begin{abstract}
Generating an articulated object from images and text requires more than plausible parts and movable joints: the assembled mechanism must execute the intended motion. This is particularly challenging for closed-chain and coupled mechanisms, where an error in one connection can block motion or separate connected parts. We introduce \method{}, Mechanism-Aware Reflective Correction, which shares an executable kinematic graph across program construction, motion solving, and correction. The graph binds physical connections to local attachment frames and represents loop closures and joint-coordinate couplings. Given prescribed driver inputs, a numerical solver computes follower motion and associates errors with specific connections and configurations. The agent uses these diagnostics to probe possible causes, revise connections or geometry, and retest the mechanism along the original motion paths. On a 20-object test set, \method{} with Kimi-K3 achieves 67.8\% valid motion coverage and 60.0\% overall success, outperforming six baselines. Starting from the same initial candidates, correction improves these scores by 13.7 and 20.0 percentage points, respectively. Ablations further show the benefits of graph-based mechanism support and graph-guided repair. Project homepage: \url{https://demianmusk.github.io/MARC-Home/}.
\end{abstract}

\section{Introduction}

Articulated 3D assets are used in robot learning, manipulation planning, and interactive simulation \citep{xiang2020sapien}. Their usefulness depends on how connected parts move together. This becomes particularly challenging when connections form closed loops or joint motions depend on one another. Consider a scissor lift: a single driving input must rotate several links and raise the platform while keeping the hinged connections intact. Generating such an asset from images and text therefore requires specifying both its geometry and the relationships that coordinate its motion. A model can match the reference in a static pose yet lock, collide, or separate when driven. The central challenge is therefore to coordinate geometry and connections so that the complete mechanism can realize the requested motion.

Existing methods have made progress in generating part geometry and articulation structure \citep{liu2024cage,wang2026artllm}. Program-generating agents further express asset construction as editable code and iteratively revise it through execution, inspection, and feedback. Articraft provides an SDK for geometry and articulation authoring together with structured validation \citep{zhou2026articraft}, while Procedura and aDSL combine assembly or spatial relations with execution feedback and visual assessment to identify and correct generated defects \citep{lin2026procedura,wang2026adsl}. However, for closed-chain and coupled structures, code generation and correction require considering several interdependent connections together. For example, a blocked follower joint may be caused by misplaced attachments, an overly restrictive joint limit, or interference between parts. To choose an effective edit, the agent must connect the observed failure to the mechanism entities involved and account for how the change affects the rest of the assembly.

We introduce \method{}, Mechanism-Aware Reflective Correction, an agent harness built around an executable kinematic graph for generating articulated-object programs from images and text. Its central design shares this graph across construction, motion solving, and correction. Our mechanism-authoring interface lets the agent specify rigid bodies, physical connections, loop closures, and joint-coordinate couplings. An SDK and its usage instructions expose this interface to the agent. Each connection binds local attachment frames, or sockets, to the participating bodies. Numerical tools translate these declarations into kinematic constraints, solve follower motion for prescribed driver inputs, and report errors at the relevant bodies, joints, sockets, and configurations. The agent then probes possible causes and revises the affected connections or geometry. A change to linkage-layer spacing, for instance, can update body positions, socket depths, and pin lengths together. The revised mechanism is rerun along the original motion paths.

We evaluate \method{} on 20 articulated objects using measures of executability, structure, valid motion, coupled motion, and overall success. With Kimi-K3, \method{} achieves 67.8\% valid motion coverage and 60.0\% overall success. Correction improves these scores from 54.1\% and 40.0\% on the same initial candidates. Structural and coupled-motion correctness remain unchanged in this comparison, identifying motion validity as the main benefit of correction. Comparisons with six generators and ablations of mechanism support and repair guidance further examine how the shared graph contributes to generation and revision.

Our method makes three contributions:
\begin{itemize}
    \item An agent harness for articulated-object generation. \method{} uses an executable kinematic graph to connect program construction, motion solving, and reflective correction for coupled and closed-chain mechanisms.
    \item A shared representation for construction and motion solving. Physical connections bind local attachment frames to part geometry, while connection types and coordinate couplings specify kinematic constraints. Numerical tools solve for motion from the declared topology and associate motion anomalies with the relevant parts and connections.
    \item Graph-guided evaluation and reflective correction. The reference, kinematic graph, and execution diagnostics guide targeted probes and agent-generated repairs. This guidance helps pinpoint the topology or interface parameters responsible for a failure.
\end{itemize}

\section{Related Work}

\paragraph{Learning-based articulated object prediction and generation.}
These methods use 3D shape models to generate part geometry and learned models to predict or jointly generate articulation relationships. Joints are typically represented through predefined parameter fields describing their type, axis, and range of motion. NAP and CAGE combine graph representations with diffusion \citep{lei2023nap,liu2024cage}, MIDGArD separates graph and shape generation \citep{leboutet2024midgard}, and ArtFormer uses a token tree \citep{su2025artformer}. URDFormer and SINGAPO infer articulated assets from images \citep{chen2024urdformer,liu2025singapo}, while PhysX-Anything generates physical assets from images \citep{cao2025physxanything}. Particulate predicts articulation from static meshes, and Instruct-Particulate adds kinematic control \citep{li2025particulate,li2026instructparticulate}. ArtLLM and URDF-Anything+ generate parts and joints autoregressively; ArtLLM also corrects joint limits using physics-based feedback \citep{wang2026artllm,wu2026urdfanythingplus}. \method{} binds joint connections to body-local attachment frames, grounding each physical edge of the kinematic graph in a geometric attachment location. Topology, interface parameters, and joint-coordinate couplings then participate together in motion solving and correction.

\paragraph{Agentic 3D code generation.}
These methods assign the construction of part geometry and articulation relationships to code-generating agents. The agents create parts through procedural modeling or calls to 3D tools and declare their connections in code, with method-specific checks, feedback, and iterative correction. Real2Code reconstructs articulated objects as code \citep{mandi2024real2code}, and Articulate-Anything combines mesh retrieval with visual actor--critic refinement \citep{le2024articulate}. Articraft provides a geometry and articulation SDK \citep{zhou2026articraft}. We draw on its program representation format in designing our mechanism-authoring interface for physical connections, loop closures, and joint-coordinate couplings. Procedura uses typed mates, assembly checks, visual refinement, and simulator validation \citep{lin2026procedura}, while aDSL combines relational operators with a Plan--Execute--Critic workflow \citep{wang2026adsl}. More broadly, tool interfaces and iterative feedback have been studied in SWE-agent and Self-Refine \citep{yang2024sweagent,madaan2023selfrefine}, and 3DCodeBench evaluates procedural modeling and human preferences \citep{gao2026codebench}. \method{} organizes execution feedback around mechanism connections and constraints so that the agent can investigate and revise the entities involved in a motion failure.

\paragraph{Closed-chain kinematics and computational mechanism design.}
Rigid-body kinematics and geometric constraint solving provide tools for finding compatible configurations \citep{lynch2017modern,fudos1997graph}. Computational design systems optimize connectivity and continuous parameters to realize target motion \citep{coros2013mechanical,thomaszewski2014linkage}. LinkEdit preserves selected functional properties when editing a working planar linkage through symbolic kinematics \citep{baecher2015linkedit}. Kinematic Kitbashing takes an abstract kinematic graph and reusable articulated parts, then optimizes their placement, orientation, and scale using a kinematics-aware attachment energy that maintains geometric consistency of attachment neighborhoods throughout articulation, together with functional objectives \citep{guo2026kinematic}. Its treatment of attachment geometry during motion is an important reference for our work. \method{} applies these principles to generation from images and text by binding joint interfaces to part geometry through local attachment frames. The agent constructs the initial mechanism, and numerical execution provides diagnostics that localize topology and interface-parameter issues for subsequent program correction.

\section{Method}
\label{sec:method}

\begin{figure}[t]
\centering
\includegraphics[width=\linewidth,trim=0 28bp 0 0,clip]{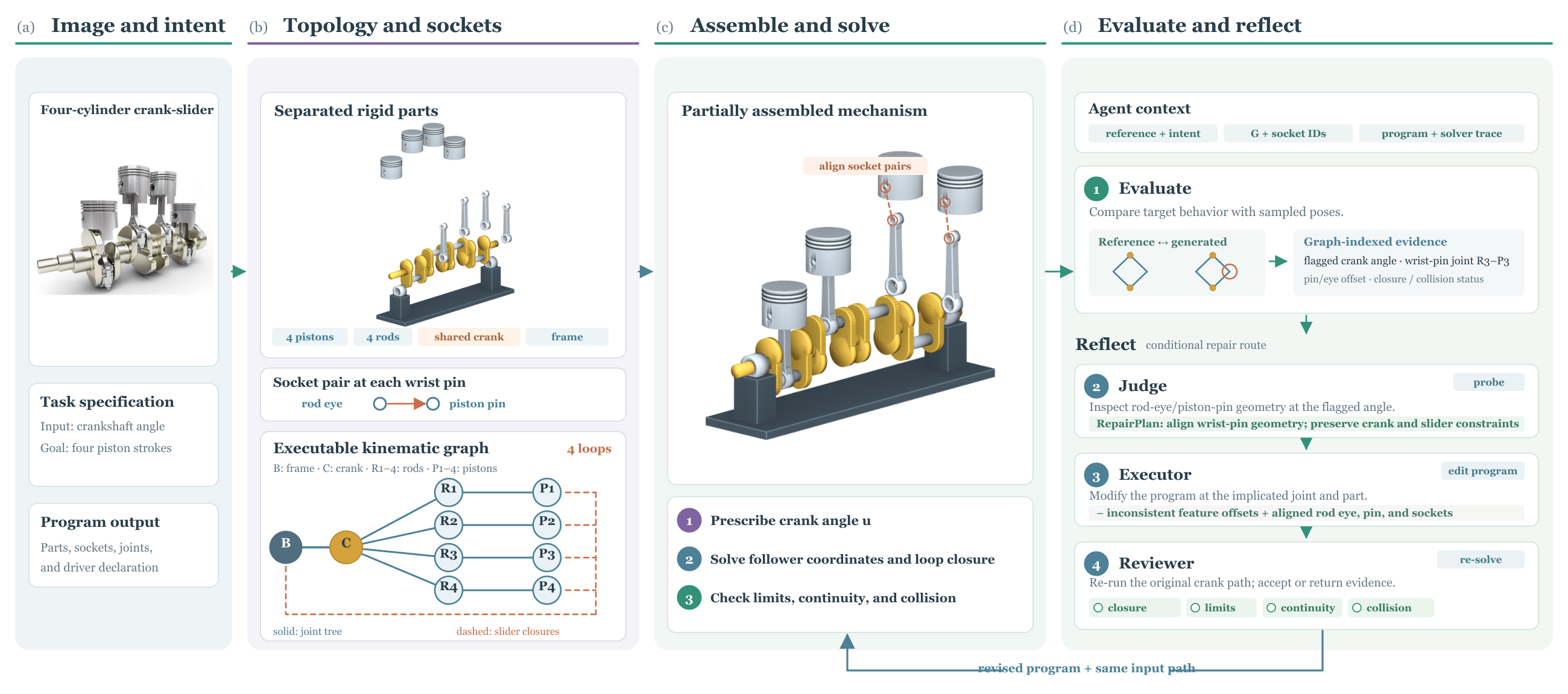}
\caption{\method{} workflow for a four-cylinder crank--slider. The agent constructs parts and socket-bound connections from the reference inputs. Solver assembles the mechanism and computes piston and connecting-rod motion under crankshaft rotation. Graph-associated diagnostics guide probing and repair, after which the revised program is rerun along the original path. The illustrated socket edit is schematic.}
\label{fig:pipeline}
\end{figure}

Given reference images $I$, a text description $x$, and a target motion domain $\mathcal U$, \method{} generates a program $P$ that constructs an articulated object matching the required appearance, connections, and motion. A prescribed driver input $u$ fixes one or more joint coordinates; the solver determines the remaining follower coordinates $q_f$ from the mechanism constraints. Motion is tested along specified paths through $\mathcal U$. In the four-cylinder crank--slider in Figure~\ref{fig:pipeline}, the driver is the crankshaft angle, and the followers include piston displacements and connecting-rod angles.

\method{} is an agent harness built around an executable kinematic graph. It equips a multimodal LLM with a mechanism-authoring SDK, numerical tools, and an evaluation-and-correction workflow. The agent writes connection and coupling declarations, interprets execution results, and chooses edits. Numerical tools construct the constraints, solve for motion, and measure violations. The harness maintains the current program, execution state, and diagnostics. The kinematic graph connects these decisions to the same rigid bodies, connections, and sockets throughout the workflow.

As shown in Figure~\ref{fig:pipeline}, Topology generates part-geometry code and code that binds inter-part joints to local attachment interfaces from the reference inputs. Solver assembles it and computes motion, and Evaluate compares the results and rendered views with the reference requirements. If correction is needed, the Reflect stage comprises three agent roles: Judge investigates possible causes through targeted probes and produces a repair plan, Executor implements the plan, and Reviewer checks the edits. The revised program returns to Solver and Evaluate along the original motion paths. The number of correction rounds is bounded by a preset limit. Sections~\ref{sec:representation}--\ref{sec:evaluate} describe the representation, numerical execution, and correction process.

\subsection{An Executable Kinematic Graph for Mechanism Authoring}
\label{sec:representation}

The representation must describe where bodies connect and how they can move relative to one another, while remaining as general as possible to support complex mechanisms. Articraft provides a URDF-based interface for parts, joints, and mimic relationships \citep{zhou2026articraft}. Drawing on its program representation format, we design a mechanism-authoring interface for the executable kinematic graph, describing mechanisms through socket-bound physical connections, loop closures, and joint-coordinate couplings. The SDK and its usage instructions expose this interface to the agent, enabling it to author closed chains and coordinate edits to attachments and geometry.

\paragraph{Connections and kinematic constraints.}
The kinematic graph is
\begin{equation}
    G = (V, E_T \cup E_L),
    \label{eq:kinematic-graph}
\end{equation}
In Equation~\ref{eq:kinematic-graph}, $V$ contains rigid bodies, $E_T$ is the joint tree used to compute body poses recursively, and $E_L$ contains additional physical connections that close loops. Each body carries the geometry of its constituent parts; graph nodes denote rigid bodies, not individual geometric primitives or auxiliary software links. From the agent's SDK declarations, numerical tools construct the kinematic constraint set $\mathcal A$ for all declared physical connections and explicit coordinate couplings. The graph $G$ records connectivity and connection attributes, while the constraint set $\mathcal A$ specifies the conditions that valid configurations must satisfy. Dependencies between joint coordinates follow from these constraints, including those induced implicitly by loop closure.

\begin{figure}[t]
\centering
\includegraphics[width=\linewidth]{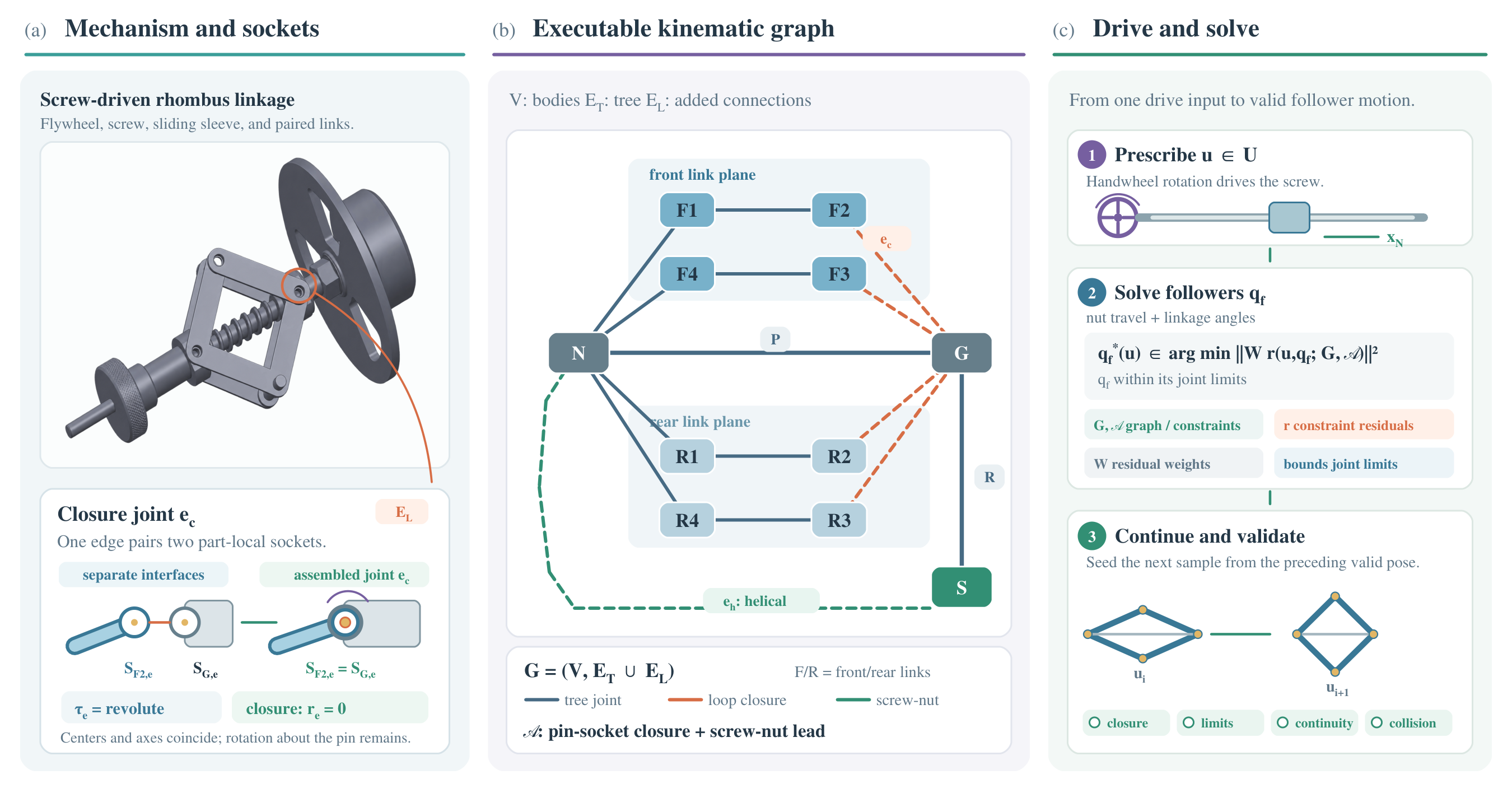}
\caption{Socket-bound connections in a screw-driven rhombus linkage. (a) Local attachment frames define the revolute closure. (b) Tree joints, loop closures, and the screw--nut connection define the kinematic graph. (c) Flywheel rotation drives nut translation and link rotation; sampled configurations are checked for closure, joint limits, continuity, and collision.}
\label{fig:kinematic-mechanism}
\end{figure}

\paragraph{Binding connections to geometry through sockets.}
Each physical connection edge $e=(a,b)$ binds two \emph{body-local sockets}, $S_{ae}$ and $S_{be}$, and specifies a connection type $\tau_e$. A socket is an attachment frame fixed to its rigid body: the paired sockets locate the connection in the geometry, while $\tau_e$ determines the permitted relative motion. Fixed, revolute, prismatic, spherical, and helical connections impose different constraints on these frames. For example, the revolute closure in Figure~\ref{fig:kinematic-mechanism}(a) aligns the pin centers and axes while allowing rotation about the common axis. The off-tree screw--nut connection in Figure~\ref{fig:kinematic-mechanism}(b) belongs to $E_L$; it aligns the screw and nut axes and couples relative rotation to axial displacement according to the screw lead. Both its attachment and coupling constraints are included in $\mathcal A$.

Sockets and geometric features share construction parameters. Changing the spacing between linkage layers can therefore update body positions, socket depths, and pin lengths together to improve clearance while preserving the connections. This lets the agent revise geometric clearance and the associated attachments as a coordinated edit.

\subsection{Motion Solving and Diagnostics from the Kinematic Graph}
\label{sec:execution}

Solver executes the current program using its graph $G$, socket frames, geometry, and constraint set $\mathcal A$, together with the target motion domain $\mathcal U$ and prescribed paths. It assembles the mechanism, solves follower motion, checks validity, and returns diagnostics associated with the mechanism.

\paragraph{Assembling the initial mechanism.}
Solver first seeks a configuration that satisfies the declared connections and explicit coordinate couplings. The joint tree computes body poses from joint coordinates, while off-tree connections constrain the relative poses of their socket frames. The initial solve satisfies these conditions together.

\paragraph{Solving for follower coordinates.}
For each driver input $u\in\mathcal U$, Solver fixes the driver coordinates and solves for the unknown follower coordinates $q_f$. In the screw-driven linkage, the screw--nut constraint relates handwheel rotation to nut travel, and loop closure determines the connected links' rotations. The residual vector $r(u,q_f;G,\mathcal A)$ collects the constraints not already enforced by the joint-tree parameterization or explicit coordinate substitution. Equation~\ref{eq:solve} computes the follower coordinates by minimizing their weighted residuals within the joint limits:
\begin{equation}
    q_f^\star(u) \in
    \arg\min_{q_f \in [\underline q_f,\overline q_f]}
    \left\|
        W\,r(u,q_f;G,\mathcal A)
    \right\|_2^2,
    \label{eq:solve}
\end{equation}
In Equation~\ref{eq:solve}, the bounds on $q_f$ specify the limits of the unknown follower coordinates, and $W$ normalizes the residual components. Driver values are sampled along the prescribed paths through $\mathcal U$. Each solved configuration initializes the next sample to continue along the same motion branch.

\paragraph{Checking motion validity.}
At each sampled configuration, Solver checks constraint residuals against their tolerances, joint coordinates against their limits, continuity between successive configurations, and prohibited geometric interference. A solution is accepted when it passes these checks. Motion validation follows the prescribed samples.

\paragraph{Associating diagnostics with the mechanism.}
Solver returns motion results and diagnostics associated with the current program state, driver values, and named bodies, joints, or sockets. A joint at its limit, an excessive closure residual, or a collision identifies a specific configuration and set of entities for the agent to inspect. These records provide the evidence used in correction.

\subsection{Graph-Guided Evaluation and Reflective Correction}
\label{sec:evaluate}
\label{sec:plan}

Existing 3D agents refine generated assets through visual comparison, programmatic checks, or simulation feedback \citep{le2024articulate,zhou2026articraft,wang2026adsl,lin2026procedura}. In a closed-chain mechanism, however, a failure can involve several interdependent connections, so an observed discrepancy alone does not determine what should change. \method{} uses the kinematic graph $G$, constraints $\mathcal A$, socket frames, and Solver diagnostics throughout evaluation and correction. They guide the agent in locating issues, investigating possible causes, and deciding which connections or parameters to revise while preserving the required motion.

\paragraph{Evaluate: identifying issues in the graph and geometry.}
Evaluate compares rendered views and motion results with the reference requirements. It checks required parts and physical connections, inspects the constraints relating driver and follower coordinates, and uses Solver diagnostics to locate execution failures. Each finding is associated with the relevant bodies, joints, or sockets and the configuration where it occurs. The assessment covers both the reference requirements and the constraints declared in the current program.

\paragraph{Judge: probing constraints and planning repairs.}
Within the Reflect stage, Judge uses the implicated graph edges and constraints to select geometric or kinematic probes at the reported failure configurations. For example, if closure fails when a follower reaches its limit, Judge can inspect the associated socket alignment and closure residuals, then temporarily relax the limit in a probe to test whether it restricts closure. Combining the probe results with the reference, Judge produces a repair plan specifying the parts, connections, or socket parameters to modify and the required connections and coordinate couplings to preserve. The graph therefore guides both probe selection and the scope of the repair; an issue at one connection is assessed in relation to the rest of the mechanism.

\paragraph{Executor and Reviewer: implementing and checking edits.}
Executor applies the plan, and Reviewer checks whether the requested changes were made and the required connections and coordinate couplings were preserved. The revised program returns to the Solver, which reconstructs its constraints and tests the complete mechanism along the original paths in $\mathcal U$. Evaluate then checks the new results against the reference. This repetition exposes both local improvements and failures elsewhere in the mechanism. Unresolved issues guide subsequent correction within the preset maximum number of correction rounds.

\section{Experiments}
\label{sec:experiments}
Our evaluation addresses three questions: whether \method\ improves the generation of complex coupled mechanisms relative to existing methods; whether its benefits transfer across foundation models; and which components contribute to structural and motion correctness.

\subsection{Experimental Setup}
\label{sec:protocol}

\paragraph{Test set and baselines.}
We evaluate on a 20-object test set of complex articulated mechanisms spanning scissor and parallelogram linkages, multi-link closed chains, rotary--sliding couplings, gear transmissions, and folding structures. Each object has reference images and a standardized task description specifying the driver and intended motion. Human annotations identify critical part roles, required connections, and expected follower responses. Outputs are judged against these requirements rather than an exact reference part decomposition. We compare \method{} with six methods: PhysX-Anything, URDF-Anything+, ArtLLM, Instruct-Particulate, Procedura, and Articraft~\citep{cao2025physxanything,wu2026urdfanythingplus,wang2026artllm,li2026instructparticulate,lin2026procedura,zhou2026articraft}. Each method generates and executes objects in its native representation. We provide matched reference information where interfaces permit and record required input or format adaptations. Articraft, the closest agent-based baseline, uses the same backbone model as \method{} in the paired comparison. A common evaluation procedure scores all outputs independently of generation; \method{}'s internal evaluation and reflection do not determine its final scores.

\paragraph{Mechanism metrics.}
We report executable rate (ER), structural correctness (SC), valid motion coverage (VMC), coupled-motion correctness (CMC), and overall success rate (OSR). ER measures whether an output loads and exposes a usable motion-control interface. SC measures satisfaction of annotated part-role and connection requirements. VMC measures the fraction of prescribed samples reached from the initial state along a valid branch that satisfies connection, joint-limit, and collision criteria. CMC assesses follower participation, direction, phase, and travel. OSR requires the critical structural and motion requirements and complete valid sampled coverage. Missing outputs and unreachable samples count as failures. For VMC, path scores are averaged within each object, then equally across objects. The remaining scores are computed within each object and averaged equally across objects; CMC uses objects with coupled-motion requirements. A stationary output fails a motion task.

\paragraph{Geometric agreement.}
On the same 20 objects, we compare \method{} and Articraft with manually constructed reference models using static gIoU, mIoU, and part-surface symmetric mean-squared Chamfer distance PC, together with motion gIoU, PC, and whole-object symmetric mean-squared Chamfer distance OC. Driver inputs are approximately matched. Functional part correspondences remain fixed across states, and missing critical parts count as errors. Geometric scores are aggregated within each object and averaged equally across objects. Definitions and aggregation details appear in Appendix~\ref{app:protocol}.

\subsection{Comparison with Existing Methods}

\begin{table}[H]
\centering\small
\setlength{\tabcolsep}{4pt}
\caption{Mechanism metrics on 20 objects (\%; 95\% intervals in Appendix~\ref{app:metric-ci}).}
\label{tab:main}
\begin{tabular}{lrrrrr}
\toprule
Method & ER & SC & VMC & CMC & OSR \\
\midrule
PhysX-Anything & 25.0 & 3.5 & 0.0 & 0.0 & 0.0 \\
URDF-Anything+ & 70.0 & 5.2 & 0.0 & 0.0 & 0.0 \\
ArtLLM & 80.0 & 27.3 & 0.0 & 0.0 & 0.0 \\
Instruct-Particulate & 55.0 & 38.0 & 0.0 & 0.0 & 0.0 \\
Procedura & 90.0 & 70.5 & 6.4 & 10.5 & 0.0 \\
Articraft (Kimi-K3) & 100.0 & 62.5 & 0.0 & 24.6 & 0.0 \\
\method\ (Kimi-K3) & \textbf{100.0} & \textbf{100.0} & \textbf{67.8} & \textbf{98.2} & \textbf{60.0} \\
\bottomrule
\end{tabular}
\end{table}

The first four baselines in Table~\ref{tab:main} produce executable outputs for some objects but satisfy few critical part-role and connection requirements, and none completes the prescribed mechanism tasks. Procedura and Articraft improve executability or structure, with limited valid motion coverage. \method{} improves motion validity most clearly in the same-backbone comparison with Articraft. Both methods achieve 100\% executability, but \method{} reaches 67.8\% valid motion coverage, compared with 0\% for Articraft. Articraft attains 24.6\% coupled-motion correctness: some followers respond as intended, although their trajectories violate connection, limit, or collision requirements. In the dual-head lamp example, the heads respond to the input while the crossing pivots separate. \method{}'s shared representation supports checking and revising these mechanism relationships throughout generation.

\method\ achieves 100\% structural correctness and 98.2\% coupled-motion correctness, whereas its valid motion coverage and overall success rate are 67.8\% and 60.0\%, respectively. This gap shows that satisfying structural requirements and producing the intended follower response does not guarantee a valid trajectory: closure errors and geometric interference can still cause failure. For example, the folding desk encounters interference before completing its prescribed travel, and the excavator fails the bucket-closure check. These cases identify motion validity and geometric clearance as remaining limitations.

\method{} also improves all six aggregate geometric-agreement metrics in Table~\ref{tab:proxy_geometry}. The object-level results reveal a useful distinction between geometric overlap and mechanism function. Articraft achieves higher static gIoU and mIoU on the gear-train example, but its transmission relationship is incorrect. The mechanism and geometric metrics capture these two aspects separately.

\begin{figure}[!t]
\centering
\includegraphics[width=0.8\linewidth]{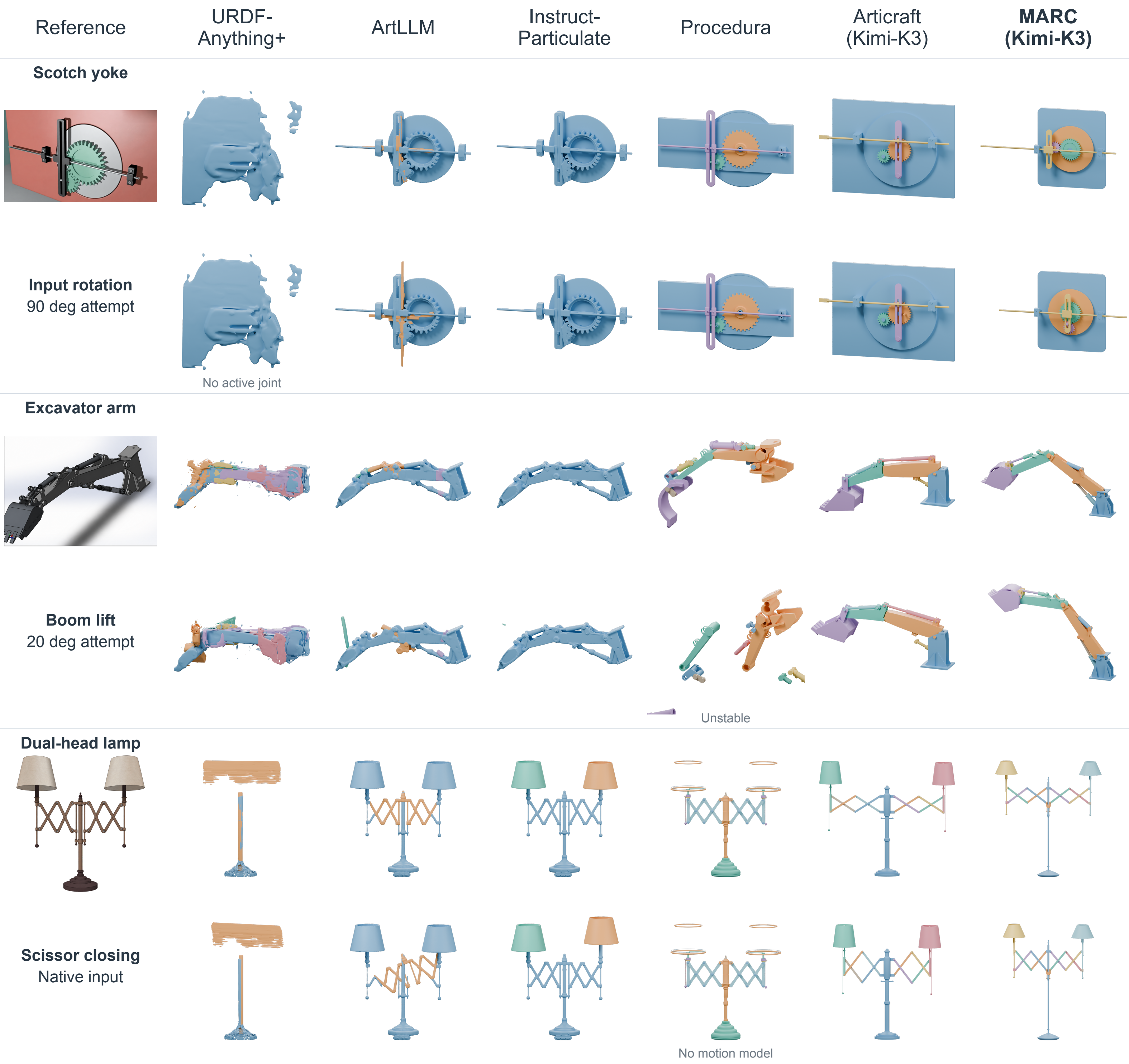}
\caption{Generated mechanisms before and after native-control inputs. Rows show input rotation, boom lifting, and scissor contraction. Each pair uses the same camera, scale, and part colors. The examples include stationary and unstable outputs; the \method{} excavator exhibits a collision.}
\label{fig:qualitative}
\end{figure}

\begin{table}[H]
\centering
\small
\setlength{\tabcolsep}{4pt}
\caption{Geometric agreement with manually constructed and annotated reference models for each object. Objects are equally weighted; motion inputs are approximately matched.}
\label{tab:proxy_geometry}
\begin{tabular}{lrrrrrr}
\toprule
& \multicolumn{3}{c}{Static}
& \multicolumn{3}{c}{Motion (approximate)} \\
\cmidrule(lr){2-4}
\cmidrule(lr){5-7}
Method
& gIoU $\uparrow$
& PC $\downarrow$
& mIoU $\uparrow$
& gIoU $\uparrow$
& PC $\downarrow$
& OC $\downarrow$ \\
\midrule
Articraft (Kimi-K3)
& $-0.0479$ & 0.0726 & 0.2316
& $-0.0497$ & 0.0743 & 0.0103 \\
\method{} (Kimi-K3)
& \textbf{0.1402} & \textbf{0.0100} & \textbf{0.3046}
& \textbf{0.0897} & \textbf{0.0130} & \textbf{0.0053} \\
\bottomrule
\end{tabular}

\end{table}

\subsection{Ablation Analysis}
We first compare direct API generation with the complete \method\ harness under the same backbone model. Direct API generates a program from the task inputs without \method's staged execution diagnostics and reflective correction; full \method\ runs the complete workflow independently from the same reference inputs. The Kimi-K3 comparison covers 20 objects, and the GPT-6 Astra comparison covers 19. Within each backbone, both configurations are evaluated on the same objects.

The main difference lies in valid motion. With Kimi-K3, Direct API achieves zero valid motion coverage and overall success, whereas full \method{} reaches 67.8\% and 60.0\%, respectively. With GPT-6 Astra, Direct API already approaches perfect structural and coupled-motion correctness, but its overall success rate is 31.6\%; full \method{} succeeds on all 19 tested objects. Thus, even when a backbone produces largely correct part relationships and follower responses, its output may fail to complete the prescribed sampled motion paths while satisfying connection and collision requirements. The largest gains are in completing the prescribed motion while maintaining the required connections and avoiding prohibited collisions.

This comparison evaluates the complete workflow, including execution feedback, tool calls, and revision. Table~\ref{tab:ablation} then examines the contributions of correction, mechanism support, and repair guidance.

\begin{table}[!tbp]
\centering\small
\setlength{\tabcolsep}{2.5pt}
\caption{Backbone and harness comparison (\%; 95\% intervals in Appendix~\ref{app:metric-ci}).}
\label{tab:backbone}
\begin{tabular}{llrrrrr}
\toprule
Backbone & Configuration & ER & SC & VMC & CMC & OSR  \\
\midrule
Kimi-K3 & Direct API & 80.0 & 54.0 & 0.0 & 61.4 & 0.0  \\
 & \method & 100.0 & 100.0 & 67.8 & 98.2 & 60.0 \\
GPT-6 Astra & Direct API & 100.0 & 97.2 & 33.5 & 98.1 & 31.6  \\
 & \method & 100.0 & 100.0 & 100.0 & 100.0 & 100.0  \\
\bottomrule
\end{tabular}
\end{table}

Table~\ref{tab:ablation} examines three configurations. Without reflection evaluates the initial Solver candidate. Without graph-based mechanism support regenerates the program using the direct-motion interface in Appendix $\mathcal A$.7, removing the graph, its associated solver and construction interfaces, and the supplementary SDK guidance. Without graph-guided repair retains graph-based execution but withholds connection localization and constraint associations from correction.

Reflection increases valid motion coverage from 54.1\% to 67.8\% and overall success from 40.0\% to 60.0\%, while structural and coupled-motion correctness remain unchanged. Its main contribution in this comparison is therefore improved motion validity. Removing graph-based mechanism support reduces coupled-motion correctness from 98.2\% to 45.8\% and valid motion coverage to 34.4\%. Removing graph-guided repair also lowers all four reported metrics. Together, these results show the value of the graph-based authoring and execution support and of using its connection and constraint information during correction.

\begin{table}[!tbp]
\centering\small
\caption{Ablation results under independent evaluation. The no-graph configuration is regenerated using the execution setup described in Appendix~\ref{app:no-graph}.}
\label{tab:ablation}
\begin{tabular}{lrrrr}
\toprule
Configuration & SC & VMC & CMC & OSR \\
\midrule
Full \method & 100.0 & 67.8 & 98.2 & 60.0 \\
Without reflection & 100.0 & 54.1 & 98.2 & 40.0 \\
Without graph-based mechanism support & 73.5 & 34.4 & 45.8 & 30.0 \\
Without graph-guided repair & 85.7 & 56.3 & 

82.9 & 45.0 \\
\bottomrule
\end{tabular}
\end{table}

Each post-repair candidate receives an independent final assessment of structure and motion. This assessment records repaired failures and regressions in properties that were initially correct.

\section{Conclusion}
We introduced \method{} for generating articulated-object programs from images and text. Its shared kinematic graph connects geometry and physical relationships to numerical motion solving and agent-guided correction. By associating execution diagnostics with specific mechanism entities, \method{} helps the agent probe motion failures and coordinate edits to connections and geometry. On the 20-object Kimi-K3 evaluation, correction primarily improves valid motion, raising coverage from 54.1\% to 67.8\% and overall success from 40.0\% to 60.0\%. Adaptive probe selection and reuse of numerical results are promising directions for reducing the computational cost of correction.

\bibliography{marc_references}
\bibliographystyle{iclr2027_conference}

\clearpage
\appendix
\section{Implementation Details}
\label{app:method}

This appendix supplements the program interfaces, numerical execution, evidence organization, and reflective correction workflow of \method{}. Following the notation in the main text, the inputs comprise reference images $I$, an object and motion description $x$, and a target motion domain $\mathcal U$ with prescribed test paths. The output is an executable articulated-object program $P$. An executable kinematic graph associates geometry, physical connections, motion constraints, and repair operations throughout the workflow.

\begin{figure}[!tbp]
    \centering
    \includegraphics[width=\linewidth]{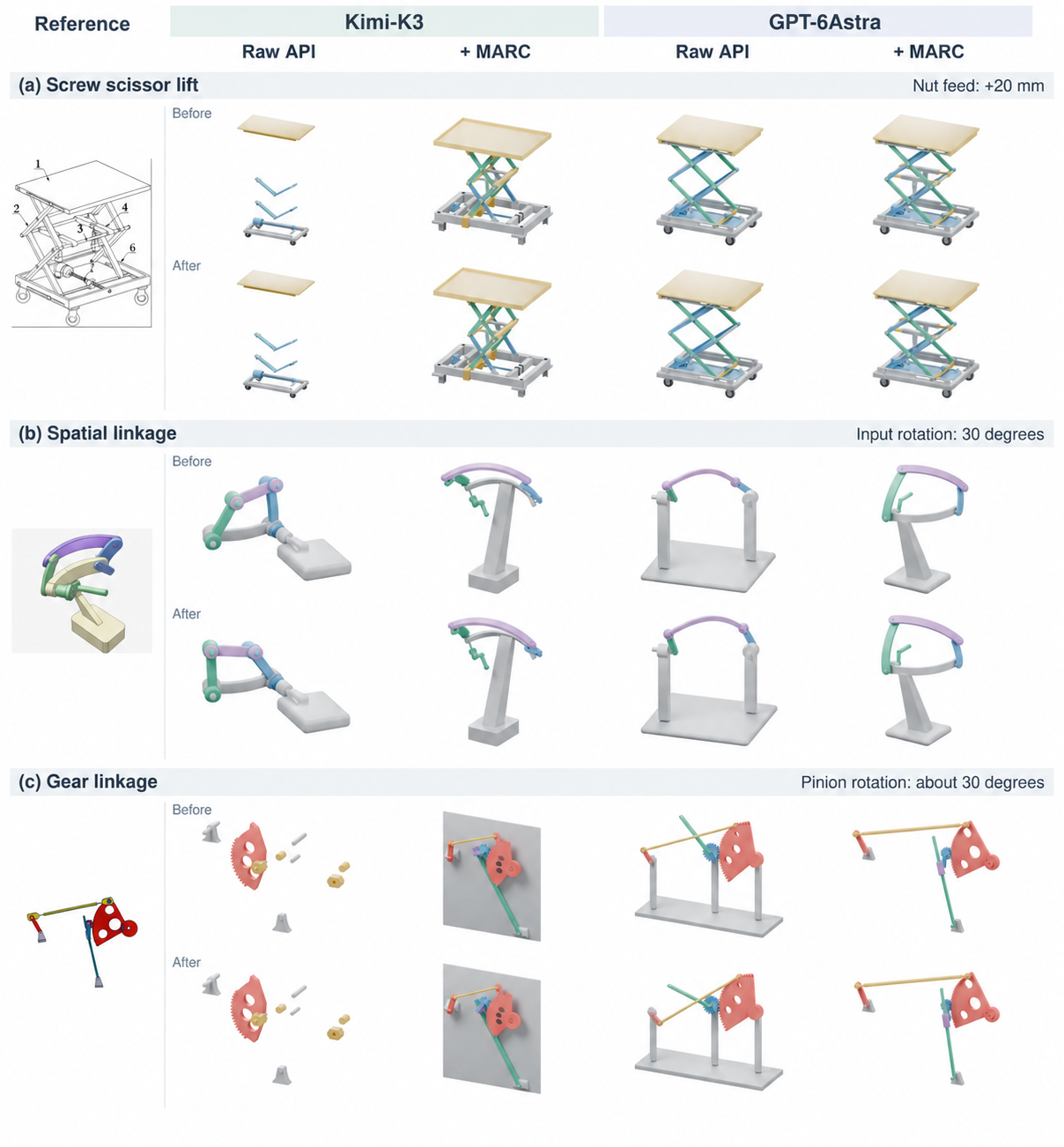}
    \caption{
    \textbf{Ablation 1: direct API generation and the full MARC
    harness across backbone models.}
    Within each backbone, direct API generation and the complete
    MARC harness use the same reference inputs.
    The comparison evaluates the system-level effect of the full
    harness, including tool calls, execution feedback, and
    reflective correction, rather than isolating a single
    component.
    The visualization illustrates the generated results;
    structural and motion metrics are computed according to
    the evaluation protocol in this appendix.
    }
    \label{fig:api-ablation-comparison}
\end{figure}

\subsection{Programming Environment and Workflow}
\label{app:workflow}

\method{} is implemented in Python and extends Articraft's programmatic authoring interface with mechanism construction and diagnostic capabilities. Geometry is constructed using CadQuery, while numerical computation uses NumPy and SciPy. Lengths and angles are expressed in meters and radians, respectively. A generated program declares geometry, rigid-body membership, local attachment frames, joints, loop-closing connections, and coordinate couplings. The numerical backend constructs and executes the mechanism from these declarations.

The workflow comprises Topology, Solver, Evaluate, and Reflect. Reflect contains three roles: Judge, Executor, and Reviewer. Language-model roles have distinct task instructions, tool interfaces, and contexts; Solver performs numerical assembly and motion computation.

\begin{table}[t]
\centering\small
\caption{Responsibilities and outputs of the workflow stages.}
\label{tab:implementation-roles}
\begin{tabularx}{\linewidth}{lXX}
\toprule
Stage or role & Main responsibility & Output \\
\midrule
Topology &
Write a mechanism program from the reference inputs using SDK documentation, construction interfaces, and examples. &
Geometry, connection, and coupling declarations. \\
Solver &
Assemble the mechanism, compute follower motion, and check constraints. &
Execution state, configurations, and numerical diagnostics. \\
Evaluate &
Compare structure, motion, and geometry with the reference requirements. &
Findings associated with mechanism entities. \\
Judge &
Investigate causes using targeted probes and plan corrections. &
A structured repair plan. \\
Executor &
Implement the plan and provide execution evidence. &
A revised candidate program. \\
Reviewer &
Check implementation of the plan and preservation of required relationships. &
A review decision and remaining items. \\
\bottomrule
\end{tabularx}
\end{table}

The Kimi-K3 configuration uses the high reasoning setting. Role specialization is implemented through inference-time context and tool access, without separately training a model for each role.

\subsection{Kinematic Graph and Geometry Interfaces}
\label{app:graph-interfaces}

The executable kinematic graph is
\begin{equation}
    G=(V,E_T\cup E_L),
\end{equation}
where $V$ denotes rigid bodies, $E_T$ is the joint tree used for forward kinematics, and $E_L$ contains additional physical connections that close mechanism loops. A rigid body may contain multiple geometric components and participate in multiple loops through different connections. Geometric primitives and auxiliary coordinate carriers used to represent compound joints are not counted as separate physical bodies.

A physical connection $e=(a,b)$ binds two body-local sockets, $S_{ae}$ and $S_{be}$, and specifies a connection type $\tau_e$. Given body transforms $T_a(q)$ and $T_b(q)$, the corresponding attachment frames in world coordinates are
\begin{equation}
    T_a(q)S_{ae},
    \qquad
    T_b(q)S_{be}.
\end{equation}
The numerical backend constrains their relative position and orientation according to the connection type.

\begin{table}[t]
\centering\small
\caption{Physical connections and coordinate couplings supported by the mechanism interface.}
\label{tab:implementation-connections}
\begin{tabularx}{\linewidth}{lXX}
\toprule
Relationship & Constraints & Remaining relative motion \\
\midrule
Fixed connection &
Position and orientation agreement. &
None. \\
Revolute connection &
Coincident attachment centers and aligned axes. &
Rotation about the common axis. \\
Prismatic connection &
Transverse position and orientation agreement. &
Translation along the guide axis. \\
Spherical connection &
Coincident attachment centers. &
Relative rotation. \\
Helical connection &
Aligned screw axes and coupled axial displacement and rotation. &
Coupled translation and rotation. \\
Affine coordinate coupling &
$q_b=\alpha q_a+\beta$. &
The dependent coordinate is determined by the driving coordinate. \\
\bottomrule
\end{tabularx}
\end{table}

For a helical connection, axial displacement and relative rotation satisfy
\begin{equation}
    d=d_0+h\theta,
\end{equation}
where $h$ is the signed axial travel per radian. Physical connection constraints and explicit coordinate couplings jointly form the constraint set $\mathcal A$ defined in the main text. Loop closure can additionally determine nonlinear follower relationships implicitly through geometry.

Geometric features and sockets share construction parameters. For example, named points on a rigid member determine both hole locations and attachment positions. Layer parameters jointly determine plate thicknesses, clearances, socket depths, and pin spans. The SDK provides composable interfaces for rigid members, planar construction, pin stacks, scissor cells, telescopic actuators, and affine drives, while retaining low-level geometry and connection declarations. These interfaces allow coordinated edits to geometry and attachment parameters.

\paragraph{SDK guidance for common joints and connections.}
Beyond the basic authoring interface, we supplement the SDK with implementation guidance and construction examples for common joints and connection structures. The guidance explains how to declare solid geometry, local attachment frames, and motion constraints together.

\begin{table}[t]
\centering\small
\caption{Supplementary SDK guidance for common mechanical connections.}
\label{tab:sdk-connection-guidance}
\begin{tabularx}{\linewidth}{lXX}
\toprule
Connection & Geometry and interface guidance & Motion relationship \\
\midrule
Ball joint &
Construct the ball and receiving structure, place both sockets at the ball center, and provide rotational clearance. &
Constrain the centers to coincide while allowing relative rotation. \\
Screw--nut connection &
Declare the screw, nut, coaxial interfaces, and supporting or guiding geometry; use consistent lead, axis, and rotation conventions. &
Relate axial displacement to relative rotation and solve jointly with the remaining mechanism constraints. \\
Bolted fastening and pin joints &
Coordinate holes, shaft diameters and lengths, plate thicknesses, and layer clearances; specify rigid-body membership of the fastener. &
Maintain a fixed relative pose for fastening, or explicitly declare a revolute connection when the shaft acts as a pivot. \\
Revolute and sliding connections &
Align the rotation or guide axes, declare attachment positions and motion ranges, and construct the associated mounting or guiding geometry. &
Retain rotation about the axis or translation along it. \\
\bottomrule
\end{tabularx}
\end{table}

For a screw with signed lead $p$, the axial travel per radian is $h=p/(2\pi)$, with the sign determined by the axis and rotation conventions. Bolt geometry alone does not determine the connection's degrees of freedom: clamping corresponds to a fixed relationship, whereas using the shaft as a pivot requires a revolute declaration.

These instructions and examples are supplied as SDK context to help the model express common mechanical connections as executable geometry and constraints. The model determines the part composition, attachment locations, and parameters of each object from its reference inputs.

\subsection{Assembly and Follower-Motion Solving}
\label{app:numerical-execution}

Solver constructs the joint tree, loop-closing connections, and constraint set from the current program. It first seeks an initial assembly satisfying the declared connections. For a prescribed driver input $u$, the driver coordinates remain fixed, and the unknown follower coordinates are solved within their limits. Constraints already satisfied by the joint-tree parameterization or explicit coordinate substitution are not duplicated in the residual.

Follower solving uses the bounded residual objective from the main text:
\begin{equation}
    q_f^\star(u)
    \in
    \arg\min_{q_f\in[\underline q_f,\overline q_f]}
    \left\|
        W\,r(u,q_f;G,\mathcal A)
    \right\|_2^2.
\end{equation}

The implementation uses nonlinear least squares with separate scaling for positional and directional residuals. Along a prescribed input path, nearby solved configurations initialize subsequent samples to support continuation on a motion branch. Assembly optimization uses rigid transformations; part dimensions are controlled by the program's geometry parameters.

Candidate configurations are checked for attachment errors, joint limits, branch continuity, and geometric interference. Closed-loop motion checking supports adaptive refinement, adding driver samples where follower changes are large, residuals increase, or near-singular behavior is detected. The sampled inputs and check results are retained for each tested path.

\begin{table}[t]
\centering\small
\caption{Principal numerical settings of the implementation.}
\label{tab:numerical-settings}
\begin{tabularx}{\linewidth}{Xl}
\toprule
Parameter & Setting \\
\midrule
Attachment-position acceptance tolerance & $10^{-4}$ m \\
Attachment-axis acceptance tolerance & $10^{-3}$ rad \\
Default configuration snapshots for the assembly objective & 5 \\
Default surface samples per part & 192 \\
Maximum local--global assembly iterations & 30 \\
Langevin search-step budget & 300 \\
Monte Carlo sample budget & 30 \\
Numerical random seed & 0 \\
Initial samples for adaptive closed-loop scanning & 5 \\
Maximum samples for adaptive closed-loop scanning & 33 \\
Maximum adaptive refinement depth & 4 \\
\bottomrule
\end{tabularx}
\end{table}

Assembly-objective snapshots and motion-validation samples control different computational stages. Motion validity is assessed at the configurations actually tested along the prescribed paths, combining connection, limit, continuity, and collision evidence.

\subsection{Graph-Associated Evidence and Layered Evaluation}
\label{app:evaluation-evidence}

Solver associates diagnostics with the current program state, driver inputs, and named bodies, joints, or sockets. Each finding records where a problem occurs, the tested configuration, and the relevant measurement, such as a loop-closure residual, an active joint limit, or interference between two bodies. Program versions and model hashes accompany the evidence so that subsequent repairs refer to the corresponding executed candidate.

Evaluate combines reference images, task descriptions, the kinematic graph, constraint information, multiview renderings, and numerical diagnostics. Findings are organized into three levels:
\begin{itemize}
    \item \textbf{L1: structure and topology.}
    Check whether critical parts, rigid-body grouping, physical connections, and required transmission relationships agree with the reference.
    \item \textbf{L2: function and motion.}
    Check driver and follower responses, loop preservation, limits, branch continuity, and interference during motion.
    \item \textbf{L3: geometry and detail.}
    Check differences in part shapes, holes, slots, supports, and local dimensions.
\end{itemize}

Visual inspection combines overall views with local views when needed, using spatial layering and depth information to help distinguish occlusion and contact regions. Motion judgments also use executed configurations and constraint measurements. Findings are associated with relevant graph nodes, connections, and constraints, providing Judge with evidence for selecting probes and repair scopes.

\subsection{Structured Reflection and Coordinated Repair}
\label{app:structured-repair}

Judge investigates Evaluate's findings by inspecting the implicated connections, constraints, and geometry and selecting targeted probes. For example, when closure failure coincides with a follower reaching its limit, Judge can inspect attachment-position and axis errors, then temporarily modify the limit in a local probe to test whether it restricts closure. Probe results and reference requirements jointly inform the repair decision.

A repair plan specifies the target parts or connections, supporting diagnosis, repair intent, permitted scope, dependencies, relationships to preserve, expected effects, and rechecking requirements. Available repairs cover topology, attachment parameters, motion parameters, and geometry. For interdependent changes, the plan identifies the entities that must be updated together. Increasing linkage-layer clearance, for example, may require coordinated changes to body positions, socket depths, and pin lengths to preserve the intended connection.

Executor implements the plan in a candidate program and records execution evidence. Reviewer compares the implementation with the plan and verifies preservation of required connections and couplings. The implementation allows at most 24 execution-agent turns and three implementation-review rounds. These internal limits are managed separately from the outer correction budget $B_{\mathrm{corr}}$.

After implementation review, the candidate returns directly to Solver, which reconstructs its constraints and tests the complete mechanism along the original paths in the target motion domain. Evaluate then compares the new results with the reference requirements. Both local improvements and regressions elsewhere inform the next decision, and unresolved findings guide further correction within the remaining budget.

\paragraph{Coordinate-consistent diagnosis.}
Coordinate changes introduced during assembly are interpreted consistently during diagnosis. If a rest offset $\delta_j$ is absorbed into the joint origin without changing the axis convention, then
\begin{equation}
    q_j^{\mathrm{before}}
    =
    \delta_j+q_j^{\mathrm{after}}.
\end{equation}
The corresponding limits are transformed by the same offset:
\begin{equation}
    [\underline q_j,\overline q_j]^{\mathrm{after}}
    =
    [\underline q_j^{\mathrm{before}}-\delta_j,\,
     \overline q_j^{\mathrm{before}}-\delta_j].
\end{equation}
This mapping distinguishes a change of coordinate zero from a change in physical motion range, keeping limit analysis and subsequent edits consistent with the assembled mechanism.

\subsection{Execution Control and Result Recording}
\label{app:execution-records}

The complete workflow follows the main text:
\begin{enumerate}
    \item \textbf{Generation.}
    Topology writes an initial mechanism program from $(I,x,\mathcal U)$.
    \item \textbf{Solving.}
    Solver assembles the mechanism, computes motion along the prescribed driver paths, and produces diagnostics.
    \item \textbf{Evaluation.}
    Evaluate checks structure, function, and geometry against the reference requirements.
    \item \textbf{Reflective correction.}
    If correction is needed and budget remains, Judge produces a plan, Executor implements it, and Reviewer checks the changes.
    \item \textbf{Revalidation.}
    The revised program returns to Solver and subsequently to Evaluate. The process repeats for at most $B_{\mathrm{corr}}$ correction rounds.
\end{enumerate}

Stages share the current program, graph, and corresponding execution evidence while preserving version relationships between candidates. Output records include source programs, repair plans, relevant probes, solver configurations, sampled driver inputs, and diagnostic results, allowing changes to be traced across revisions.

Internal Evaluate judgments guide generation and correction. The ER, SC, VMC, CMC, and OSR values reported in the paper are computed by the independent evaluation procedure specified in the main text; internal judgments do not directly determine the final scores.

\subsection{Implementation of the No-Graph Ablation}
\label{app:no-graph}

The no-graph configuration removes the graph-based mechanism construction and execution support of the full method. This includes the executable kinematic graph, socket-bound connection and loop-closure declarations, the original graph-based assembly and Solver components, and the supplementary mechanism-construction interfaces and SDK guidance described above. This is therefore a joint ablation of the representation and its associated execution and authoring support.

The configuration regenerates programs from the task inputs. The model uses general-purpose geometry and numerical libraries to define rigid-body geometry and directly implement the mapping from driver inputs to world-space body poses. The program exposes two interfaces: \texttt{build()} returns each body's local geometry and appearance, and \texttt{motion(controls)} returns its world-space rigid transformation for the supplied inputs. Geometry remains fixed during motion, and the generated program performs the motion calculations itself.

The model context excludes the original graph-based construction documentation, mechanism-example retrieval, and supplementary guidance for connections such as ball joints, screw--nut mechanisms, and layered pin assemblies. The graph-based assembly, closed-loop follower solver, and graph-associated repair interfaces are also unavailable. Retained general-purpose tools execute the program, check rigid transformations, render multiple views, and record sampled configurations and geometric contacts.

During correction, the model uses the reference images, candidate code, renderings, and execution diagnostics to produce a free-form repair plan, revise the complete program, and execute it again. This preserves opportunities for execution-conditioned revision while removing localization and planning interfaces based on graph nodes, connections, and constraints.

This configuration regenerates its initial candidates and uses its own direct-motion execution interface. Its outputs are assessed using the independent structural and motion evaluation protocol in the main text. The comparison measures the combined contribution of the kinematic graph, its associated Solver and construction interfaces, and SDK mechanism guidance; it does not isolate the effect of graph notation alone.

\section{Reproducibility Details}
\label{app:settings}
\label{app:reproducibility}

This appendix describes how to restore an experimental configuration, rerun generation and correction, and verify the evaluation of saved artifacts. Method and numerical implementation details are given in Appendix~\ref{app:method}; metric definitions, motion checks, and baseline adapters are described in Appendix~\ref{app:protocol}. We distinguish replaying and rescoring saved artifacts from rerunning the complete generation workflow from reference inputs.

\subsection{Reproduction Units and Version Correspondence}

The unit of reproduction is an object-level experimental record associating reference inputs, run configurations, program versions, and results. A stable object identifier links the reference images, task description, driver domains, and evaluation requirements. File hashes distinguish the programs used for generation, correction, and evaluation, preventing code, diagnostics, or scores from different candidates from being combined.

\begin{table}[t]
\centering\small
\caption{Records used to reproduce and verify an experimental result.}
\label{tab:reproduction-records}
\begin{tabularx}{\linewidth}{lX}
\toprule
Record category & Purpose \\
\midrule
Inputs and task &
Identify reference images, text, driver domains, and test paths. \\
Implementation and configuration &
Identify the code version, SDK, prompts, example resources, and model-call settings. \\
Programs and repairs &
Associate the initial candidate, repair plans, code changes, and final candidate. \\
Execution and evaluation &
Associate solver settings, sampled configurations, diagnostics, exported artifacts, and object-level scores. \\
\bottomrule
\end{tabularx}
\end{table}

A correction experiment starts from its specified initial candidate and execution state; a regeneration experiment starts from reference inputs. These routes are recorded separately so that replaying or correcting an existing candidate is not counted as a new generation run.

\subsection{Environment and Model Configuration}

The implementation supports Python 3.11--3.12 and uses CadQuery, NumPy, SciPy, Trimesh, and collision-query components. Reproduction should restore the software versions and numerical configuration associated with the target experiment. Appendix~\ref{app:method} specifies the principal settings of the main method. Worker counts and compute devices are recorded as execution-environment parameters for interpreting resource use and latency.

The Kimi-K3 main-method configuration uses the high reasoning setting. The model identifier, reasoning setting, prompts, tool descriptions, SDK documentation, and example resources jointly define the model-call configuration. Internal execution-agent and review limits are configured separately from the outer correction budget, using the settings associated with the target experiment.

Numerical replay and external scoring of saved artifacts do not require new calls to the generation model. Regenerating candidates requires access to the corresponding model service. Numerical seeds control the numerical components; new calls to a remote generation model receive separate run identifiers, and their outputs are evaluated anew.

\subsection{Two Reproduction Routes}

\paragraph{Replay of saved artifacts.}
Load the specified program with its native connections, loops, couplings, and limits. Restore its driver inputs and test paths, execute the motion, and regenerate checking results. Apply the independent evaluation protocol to assess structure, valid motion, and follower responses, then recompute aggregate metrics from object-level results. Replay keeps the candidate unchanged and uses its program hash to verify correspondence with the reported artifact.

\paragraph{Rerunning the complete workflow.}
Start Topology from the specified reference inputs, execute Solver and Evaluate, and run Judge, Executor, and Reviewer within the corresponding budget. Each revised candidate returns to numerical execution and evaluation. The final candidate is scored by the independent evaluation procedure; internal model judgments do not replace the reported metrics.

For programs generated through the graph-based mechanism interface, replay preserves loop-closure and coordinate-coupling declarations. An execution that retains only the joint tree while discarding these relationships is not treated as an equivalent replay of the original mechanism.

\subsection{Starting Points for Comparisons and Ablations}

The configuration without reflection evaluates its corresponding initial Solver candidate without subsequent correction. Comparisons between correction mechanisms use the initial candidates specified by the respective experiment and retain the correspondence between inputs and candidates.

The no-graph configuration follows Appendix~\ref{app:no-graph}: it regenerates programs from reference inputs, uses a direct-motion execution interface, and removes the original graph-based Solver and associated SDK guidance. Its reproduction does not load the full method's graph-based execution or instructional resources. Outputs from different configurations are assessed using the independent evaluation protocol, without requiring identical internal representations or execution backends.

Baselines use their native outputs and execution adapters, as detailed in Appendix~\ref{app:protocol}. Reproduction adapters load and execute existing declarations; they do not add missing connections or modify candidate geometry.

\subsection{Verification and Aggregation}

Verification establishes the object and program version, driver inputs and sampling configuration, execution results, and object-level scores. Programs that cannot be loaded, missing outputs, and unreachable prescribed motion samples contribute to the corresponding failures under the evaluation protocol. Reruns and retries retain separate identifiers and provenance rather than silently overwriting the original candidate.

Reported aggregate metrics are recomputed from object-level evaluation results. Compilation, implementation review, and internal Solver checks are recorded as stage-specific outcomes, while final task conclusions follow the independent structural and motion assessment. This separates verification of generation outputs, correction effects, and the artifacts underlying the reported scores.

\section{Reference Ground Truth and Evaluation Protocol}
\label{app:protocol}

\subsection{Manually Constructed Reference Ground Truth}
\label{app:reference-ground-truth}

We manually construct and annotate reference models for 20 articulated objects.
Starting from procedural modeling code, we inspect and correct
these models against the input images, covering part geometry, rigid-body decomposition, connection
locations, joint types, transmission relationships, and spatial
relationships during motion. The reference ground truth includes
both geometric models and functional requirements for assessing
structural and motion correctness.

The reference information for each object covers three aspects.
First, it specifies the functional roles and relative arrangement
of required parts, rigid and articulated connections, and
loop-closing relationships. Second, it identifies independent
drivers, prescribed motion ranges, follower directions, phase
relationships, and required travel. Third, it specifies the
connections that must remain intact during motion, applicable
joint limits, and regions of permitted contact or prohibited
penetration.

Correspondences between generated and reference parts are
established through functional roles and connection relationships.
Identical part names or mesh decompositions are not required.
For example, multiple submeshes rigidly attached to one another
may correspond to a single reference part. Such correspondences
must not conceal missing articulated connections or incorrect
degrees of freedom. For details that cannot be uniquely determined
from the input images, evaluation is restricted to requirements
that can be judged reliably, while allowing reasonable alternative
realizations with equivalent functionality.

\subsection{Evaluation Units and Common Conventions}
\label{app:evaluation-conventions}

The reference collection contains $N=20$ objects.
For configuration $m$, let $\mathcal{D}_m$ denote its evaluated
object set and let $n_m=|\mathcal{D}_m|$.
Within each backbone, the direct API and full-harness configurations
use the same object set. The Kimi comparison covers 20 objects,
whereas the GPT-6 Astra comparison covers 19. Each metric uses the actual
evaluation set of the corresponding configuration rather than
assigning a denominator of 20 to every configuration.

For object $i$, let $A_i$ be the set of structural requirements
and $R_i$ the set of coupled-motion requirements.
Each requirement specifies the property being assessed and its
acceptance condition. Requirements essential to the basic
functionality of the mechanism are designated as critical.
The following equations omit the configuration index and operate
on one configuration's evaluation set $\mathcal{D}$, with
$n=|\mathcal{D}|$.

Motion evaluation starts from a prescribed initial state and
proceeds sequentially along specified driver paths.
Sample locations are determined by the requested motion domain;
the domain is not reduced retrospectively to match the range
that a generated model can complete. Separate test paths are
recorded for independent drivers and for forward and reverse
motion.

Generated results are loaded through their corresponding execution
interfaces while retaining their original connections, coordinate
couplings, and limits. Execution adapters do not add missing parts
or repair mechanism relationships. Successful loading, local joint
motion, validity over a complete trajectory, and semantic mechanism
correctness constitute distinct levels of evidence.
All primary metrics are aggregated within each object before
macro-averaging across objects.

\subsection{Primary Evaluation Metrics}
\label{app:primary-metrics}

\subsubsection{Executable Rate (ER)}

Executable Rate measures whether a generated result can be loaded
and exposes a usable motion-control interface.
Let $e_i\in\{0,1\}$ indicate whether object $i$ meets this condition.
Then
\begin{equation}
\mathrm{ER}
=
\frac{1}{n}
\sum_{i\in\mathcal{D}} e_i
\times 100\%.
\label{eq:appendix-er}
\end{equation}

A static mesh or rendered image alone does not satisfy this
condition. ER does not establish correct part structure,
transmission relationships, or motion behavior; these properties
are assessed by the following metrics.

\subsubsection{Structural Correctness (SC)}

Structural Correctness measures the fraction of reference
structural requirements satisfied by a generated result.
For requirement $a\in A_i$, define
\begin{equation}
s_{ia}
=
\mathbf{1}
\left[
a \text{ is satisfied}
\right],
\end{equation}
where $\mathbf{1}[\cdot]$ is the indicator function.
The object-level and aggregate scores are
\begin{align}
\mathrm{SC}_i
&=
\frac{1}{|A_i|}
\sum_{a\in A_i} s_{ia},
\\
\mathrm{SC}
&=
\frac{1}{n}
\sum_{i\in\mathcal{D}} \mathrm{SC}_i
\times 100\%.
\label{eq:appendix-sc}
\end{align}

Structural requirements assess the presence of necessary parts,
their functional roles, the identities of connected bodies,
joint types, attachment locations, and loop-closing relationships.
A visually similar model with incorrect connections does not
satisfy the corresponding requirements.
Missing required parts or connections cause their associated
requirements to fail.

\subsubsection{Valid Motion Coverage (VMC)}

Valid Motion Coverage measures the fraction of prescribed motion
samples that can be reached from the initial state along a valid
motion branch.
Suppose object $i$ has $M_i$ test paths, with $K_{ip}$ samples
on path $p$.
We set $v_{ipk}=1$ only when sample $k$ is reached from the prescribed
initial state along that path and the traversal satisfies all
applicable critical connection, loop-closure, joint-limit, and
collision conditions; otherwise, $v_{ipk}=0$.

The object-level score is
\begin{equation}
\mathrm{VMC}_i
=
\frac{1}{M_i}
\sum_{p=1}^{M_i}
\left(
\frac{1}{K_{ip}}
\sum_{k=1}^{K_{ip}} v_{ipk}
\right),
\end{equation}
and the aggregate score is
\begin{equation}
\mathrm{VMC}
=
\frac{1}{n}
\sum_{i\in\mathcal{D}} \mathrm{VMC}_i
\times 100\%.
\label{eq:appendix-vmc}
\end{equation}

After a blockage, connection separation, limit violation, or
prohibited collision, jumping to a later pose does not extend
the valid coverage of that path.
An isolated feasible pose that cannot be reached along the
prescribed path is not counted as a valid sample.
Forward and reverse motion are assessed using their respective
prescribed starts and paths.
Full sampled coverage does not constitute a proof of continuous
collision freedom between samples.

\subsubsection{Coupled-Motion Correctness (CMC)}

Coupled-Motion Correctness measures whether driver inputs produce
the required follower responses, including follower participation,
direction, phase, transmission relationships, and required travel.
For requirement $r\in R_i$, define
\begin{equation}
c_{ir}
=
\mathbf{1}
\left[
r \text{ is satisfied over its prescribed interval}
\right].
\end{equation}
The object-level score is
\begin{equation}
\mathrm{CMC}_i
=
\frac{1}{|R_i|}
\sum_{r\in R_i} c_{ir}.
\end{equation}

Let
\begin{equation}
\mathcal{D}_{C}
=
\left\{
i\in\mathcal{D}: |R_i|>0
\right\}
\end{equation}
denote the objects with coupled-motion requirements.
The aggregate score is
\begin{equation}
\mathrm{CMC}
=
\frac{1}{|\mathcal{D}_{C}|}
\sum_{i\in\mathcal{D}_{C}} \mathrm{CMC}_i
\times 100\%.
\label{eq:appendix-cmc}
\end{equation}

Observed displacement alone is insufficient to satisfy a
coupled-motion requirement.
A gear rotating in the wrong direction or with an incorrect
transmission relationship fails the corresponding requirement.
Likewise, small oscillations do not satisfy a requirement for
substantial follower travel.

Objects without coupled-motion requirements are excluded from
the CMC denominator rather than assigned a vacuous perfect score.
CMC and VMC assess response correctness and motion validity,
respectively: a correct local follower response does not establish
validity over the complete prescribed motion.

\subsubsection{Overall Success Rate (OSR)}

Overall Success Rate uses a binary object-level decision.
An object succeeds only if all of the following conditions hold:
the result is executable; all predesignated critical structural
requirements are satisfied; all applicable critical motion
requirements are satisfied; the prescribed paths have complete
valid sampled coverage; and the overall layout is consistent
with the functional structure of the reference object.

Let $o_i\in\{0,1\}$ denote this decision. Then
\begin{equation}
\mathrm{OSR}
=
\frac{1}{n}
\sum_{i\in\mathcal{D}} o_i
\times 100\%.
\label{eq:appendix-osr}
\end{equation}

OSR is not an arithmetic average of the other metrics.
A model that satisfies many local requirements but lacks a
critical connection, has an incorrect critical transmission,
or cannot complete the prescribed motion is not counted as an
overall success.
Successful termination of the internal Solver does not by itself
establish acceptance under the final evaluation protocol.

\subsection{Geometric Agreement Metrics}
\label{app:geometry-metrics}

\subsubsection{Correspondence, Alignment, and Normalization}

Geometric metrics compare the shape and spatial arrangement of a
generated model with its manually constructed and annotated reference model.
They are reported separately from structural and motion metrics.
Part correspondences are established by functional role, and
fixed submeshes belonging to the same rigid body may be merged
into one comparison unit.
These correspondences remain fixed throughout a motion sequence.

Geometry is compared after a single global coordinate and scale
alignment. Distances are normalized by the reference model's
bounding-box diagonal in the designated reference configuration.
Dynamic comparisons retain the same alignment transform.
Individual parts or frames are not realigned independently,
because doing so could remove genuine layout or motion errors.

\subsubsection{Part Bounding-Box Agreement: mIoU and gIoU}

Let $B$ and $B^{*}$ be the aligned three-dimensional axis-aligned
bounding boxes of a generated part and its reference counterpart.
Their intersection over union is
\begin{equation}
\mathrm{IoU}(B,B^{*})
=
\frac{
\operatorname{Vol}(B\cap B^{*})
}{
\operatorname{Vol}(B\cup B^{*})
}.
\end{equation}

The object-level mean IoU is
\begin{equation}
\mathrm{mIoU}_i
=
\frac{1}{J_i}
\sum_{j=1}^{J_i}
\mathrm{IoU}(B_{ij},B^{*}_{ij}),
\end{equation}
where $J_i$ is the number of reference comparison units.

Let $C$ be the smallest axis-aligned bounding box enclosing both
$B$ and $B^{*}$. Generalized IoU is
\begin{equation}
\mathrm{gIoU}(B,B^{*})
=
\mathrm{IoU}(B,B^{*})
-
\frac{
\operatorname{Vol}(C)
-
\operatorname{Vol}(B\cup B^{*})
}{
\operatorname{Vol}(C)
}.
\end{equation}

Object-level gIoU is likewise averaged over parts.
Higher mIoU and gIoU indicate closer agreement in spatial position
and extent.
These are three-dimensional part bounding-box metrics, not
pixel-overlap metrics computed from rendered images.

\subsubsection{Surface Distances: PC and OC}

Point clouds are sampled from the normalized mesh surfaces.
For two nonempty point sets $P$ and $Q$, we use the symmetric
mean-squared Chamfer distance:
\begin{equation}
\begin{split}
d_{\mathrm{CD}}(P,Q)
={}&
\frac{1}{2|P|}
\sum_{p\in P}
\min_{q\in Q}\|p-q\|_2^2
\\
&+
\frac{1}{2|Q|}
\sum_{q\in Q}
\min_{p\in P}\|q-p\|_2^2.
\end{split}
\end{equation}

Part-level surface distance, PC, is computed for corresponding
parts and then averaged within each object:
\begin{equation}
\mathrm{PC}_i
=
\frac{1}{J_i}
\sum_{j=1}^{J_i}
d_{\mathrm{CD}}(P_{ij},P^{*}_{ij}).
\end{equation}

Object-level surface distance, OC, compares point clouds sampled
from the complete object surfaces:
\begin{equation}
\mathrm{OC}_i
=
d_{\mathrm{CD}}
\left(
P_i^{\mathrm{obj}},
P_i^{*\mathrm{obj}}
\right).
\end{equation}

Lower PC and OC are better.
PC emphasizes the shape and location of corresponding parts,
whereas OC measures the discrepancy between complete object
surfaces.
Surface sampling uses common sampling rules and random seeds.
Missing reference parts receive a common missing-part penalty;
they are not removed from the comparison set before averaging
over the remaining parts.

\subsubsection{Static and Dynamic Comparisons}

Static comparisons use corresponding reference configurations.
Dynamic comparisons evaluate generated and reference models
at corresponding driver inputs.
For a per-state geometric metric $g$, define
\begin{equation}
g_i^{\mathrm{dyn}}
=
\frac{1}{K_i}
\sum_{k=1}^{K_i}
g\left(
X_i(u_{ik}),
X_i^{*}(u_{ik})
\right),
\end{equation}
where $X_i(u_{ik})$ and $X_i^{*}(u_{ik})$ denote the generated and
reference states at the corresponding input.

When multiple paths are evaluated, scores are first averaged
within each path and then across paths.
State correspondence is established using actual driver
coordinates or physical travel rather than frame indices alone.
When discrete recordings provide only nearby inputs, the
input mismatch is recorded and the pairing is treated as
approximate.
Non-executable results are not assigned dynamic geometric scores
by constructing substitute poses.

Geometric similarity does not replace mechanism correctness.
Models with similar shapes may still contain incorrect
connections, while functionally correct mechanisms may retain
local shape discrepancies.

\subsection{Aggregation, Failure Handling, and Confidence Intervals}
\label{app:aggregation}

\subsubsection{Aggregation and Failure Handling}

ER and OSR aggregate binary object-level outcomes.
SC, VMC, and CMC are computed per object and then macro-averaged.
Geometric metrics are likewise aggregated at the object level;
parts and motion frames are not pooled as independent
observations.

Generation failures, missing final artifacts, and results that
cannot form executable objects are treated as failures in the
primary task metrics.
For CMC, a failed object remains in the denominator whenever
it belongs to the set with coupled-motion requirements.
Infrastructure interruptions are recorded separately from
generated-program failures and handled under a common retry
policy. Failed objects are not removed to improve reported
performance.

Position, angle, joint-limit, collision, and follower-travel
judgments use common evaluation conventions.
Permitted contact and prohibited penetration are distinguished
using the reference requirements.
Adjacency or the presence of a connection does not automatically
exempt a pair of parts from collision assessment.

\subsubsection{95\% Confidence Intervals}
\label{app:metric-ci}

All statistical analyses use the object as the unit of analysis.
For the binary metrics ER and OSR, we use two-sided 95\% Wilson
score intervals.
Let $\hat p$ be the observed success proportion, $n$ the number
of objects, and $z=\Phi^{-1}(0.975)$, where $\Phi$ is the standard
normal cumulative distribution function.
The interval is
\begin{equation}
\frac{
\hat p+\dfrac{z^2}{2n}
\;\pm\;
z\sqrt{
\dfrac{\hat p(1-\hat p)}{n}
+
\dfrac{z^2}{4n^2}
}
}{
1+\dfrac{z^2}{n}
}.
\label{eq:appendix-wilson}
\end{equation}

For SC, VMC, and CMC, whose object-level scores lie in $[0,1]$,
we use conservative mean intervals obtained by inverting a
Chernoff--Hoeffding bound.
Let $n$ be the number of eligible objects and $\bar x$ their
mean score.
The interval consists of the values $\mu\in[0,1]$ satisfying
\begin{equation}
n\,\mathrm{KL}(\bar x\Vert\mu)
\leq
\log\frac{2}{\alpha},
\qquad
\alpha=0.05,
\label{eq:appendix-bounded-ci}
\end{equation}
where
\begin{equation}
\mathrm{KL}(p\Vert q)
=
p\log\frac{p}{q}
+
(1-p)\log\frac{1-p}{1-q}.
\end{equation}
Boundary cases are defined by continuity.
This construction retains nonzero interval width when all
observed scores are zero or one.
Interval endpoints are converted to percentages for reporting.

Each configuration uses its actual evaluation set to determine
the sample size.
Both Kimi configurations contain 20 objects, and both GPT-6 Astra
configurations contain 19.
CMC excludes the dishwasher, which has no coupled-motion
requirement, giving sample sizes of 19 and 18, respectively.

These are object-level marginal confidence intervals under an
assumption of independent objects from a comparable population.
They do not quantify repeated-generation variability or
annotation and adjudication error, and they do not provide
simultaneous coverage across multiple metrics.
The intervals do not make the curated evaluation set a random
population sample.
The means in the main text describe performance on the fixed
evaluation set; the intervals supplement them with uncertainty
under the stated statistical assumptions.

\subsubsection{Confidence Interval Results}

Table~\ref{tab:metric-ci} reports the two-sided 95\% confidence
intervals for all five primary metrics.
Entries are interval endpoints, not means or standard deviations.
The corresponding mean scores are reported in the main text.

\begin{table}[!tbp]
\centering
\footnotesize
\setlength{\tabcolsep}{3pt}
\renewcommand{\arraystretch}{1.12}
\caption{
Object-level two-sided 95\% confidence intervals.
All endpoints are percentages.
}
\label{tab:metric-ci}
\begin{tabular}{lrrrrr}
\toprule
Method & ER & SC & VMC & CMC & OSR \\
\midrule
PhysX-Anything
& [11.2, 46.9]
& [0.0, 25.9]
& [0.0, 16.8]
& [0.0, 17.6]
& [0.0, 16.1] \\

URDF-Anything+
& [48.1, 85.5]
& [0.1, 28.9]
& [0.0, 16.8]
& [0.0, 17.6]
& [0.0, 16.1] \\

ArtLLM
& [58.4, 91.9]
& [7.3, 57.3]
& [0.0, 16.8]
& [0.0, 17.6]
& [0.0, 16.1] \\

Instruct-Particulate
& [34.2, 74.2]
& [13.6, 67.7]
& [0.0, 16.8]
& [0.0, 17.6]
& [0.0, 16.1] \\

Procedura
& [69.9, 97.2]
& [40.5, 91.5]
& [0.1, 31.0]
& [0.7, 38.1]
& [0.0, 16.1] \\

Articraft (Kimi-K3)
& [83.9, 100.0]
& [32.8, 86.7]
& [0.0, 16.8]
& [5.6, 55.2]
& [0.0, 16.1] \\

\midrule
Kimi-K3 Direct API
& [58.4, 91.9]
& [25.4, 80.8]
& [0.0, 16.8]
& [31.1, 86.5]
& [0.0, 16.1] \\

MARC (Kimi-K3)
& [83.9, 100.0]
& [83.2, 100.0]
& [37.8, 90.0]
& [77.0, 100.0]
& [38.7, 78.1] \\

GPT-6 Astra Direct API
& [83.2, 100.0]
& [74.7, 100.0]
& [10.4, 64.2]
& [75.9, 100.0]
& [15.4, 54.0] \\

MARC (GPT-6 Astra)
& [83.2, 100.0]
& [82.4, 100.0]
& [82.4, 100.0]
& [81.5, 100.0]
& [83.2, 100.0] \\
\bottomrule
\end{tabular}

\par\vspace{4pt}
\begin{minipage}{\linewidth}
\footnotesize
Both GPT-6 Astra configurations use 19 objects; all other configurations
use 20. CMC excludes the dishwasher, giving denominators of 18
and 19, respectively.
\end{minipage}
\end{table}

\subsection{Ablation 1: Backbone Model and Harness Comparison}
\label{app:api-harness-comparison}

Ablation 1 compares direct API generation with the complete MARC
harness for each backbone model.
Within each backbone, the two configurations independently
generate results from the same reference inputs and use the same
object set and evaluation criteria.
Direct API generation does not include MARC's staged execution
diagnostics and reflective correction, whereas the full-harness
configuration executes the complete workflow.

This experiment measures the system-level effect of the full
harness, including tool calls, execution feedback, and reflective
correction. It does not attribute the observed difference to any
single component.
Sample counts are disclosed separately for each backbone.
Cross-backbone comparisons should distinguish results on each
backbone's complete evaluation set from results restricted to
their common object subset.

Figure~\ref{fig:api-ablation-comparison} provides a qualitative
comparison of the generated results.
Structural and motion judgments follow the evaluation protocol
defined in this appendix.


\end{document}